\documentclass{article}

\usepackage[numbers]{natbib}
\usepackage{amsmath}
\usepackage{wrapfig}
\usepackage{graphicx}
\usepackage{enumitem}
\usepackage{booktabs}
\usepackage{multirow}
\usepackage{makecell}
\usepackage[table]{xcolor}
\usepackage{boxedminipage}
\usepackage{tcolorbox}
\usepackage{tabularx}
\tcbuselibrary{breakable}

\definecolor{nmgray}{RGB}{229,229,229}
\definecolor{shadecolor}{RGB}{237,237,237}
\definecolor{darkpastelgreen}{rgb}{0.01, 0.75, 0.24}

\usepackage{pifont}

\usepackage{longtable}
\usepackage{adjustbox}
\usepackage{capt-of}
\usepackage{float}
\usepackage[colorlinks=true, urlcolor=blue]{hyperref}

\usepackage[preprint]{neurips_2025}

\usepackage[utf8]{inputenc} 
\usepackage[T1]{fontenc}    
\usepackage{hyperref}       
\usepackage{url}            
\usepackage{booktabs}       
\usepackage{amsfonts}       
\usepackage{nicefrac}       
\usepackage{microtype}      
\usepackage{xcolor}         

\title{FreePRM: Training Process Reward Models Without Ground Truth Process Labels}

\author{
Lin Sun, Chuang Liu, Xiaofeng Ma, Tao Yang, Weijia Lu, Ning Wu\\
UAES AI Lab\\ 
\texttt{\{lin.sun,chuang.liu,xiaofeng.ma,tao.yang9,weijia.lu,ning.wu\}@uaes.com}
}

\begin{document}

\maketitle

\begin{abstract}
  Recent advancements in Large Language Models (LLMs) have demonstrated that Process 
  Reward Models (PRMs) play a crucial role in enhancing model performance. 
  However, training PRMs typically requires step-level labels, either manually annotated 
  or automatically generated, which can be costly and difficult to obtain at scale.
  To address this challenge, we introduce {\bf FreePRM}, a weakly supervised framework 
  for training PRMs without access to ground-truth step-level labels.
  FreePRM first generates pseudo step-level labels based on the correctness of 
  final outcome, and then employs \textit{Buffer Probability} to eliminate impact of noise inherent 
  in pseudo labeling. Experimental results show that FreePRM achieves 
  an average F1 score of {\bf 53.0\%} on ProcessBench, outperforming fully supervised PRM 
  trained on Math-Shepherd by {\bf +24.1}\%. Compared to other 
  open-source PRMs, FreePRM outperforms upon RLHFlow-PRM-Mistral-8B (28.4\%) 
  by {\bf +24.6}\%, EurusPRM (31.3\%) by {\bf +21.7}\%, and Skywork-PRM-7B 
  (42.1\%) by {\bf +10.9}\%.
  This work introduces a new paradigm in PRM training, significantly reducing reliance 
  on costly step-level annotations while maintaining strong performance.
  Code is available at \url{https://github.com/sunlin-ai/FreePRM}.

\end{abstract}

\section{Introduction}\label{sec:intro}

\begin{wrapfigure}{r}{0.5\textwidth}
    \vspace{-25pt}
    \centering
    \includegraphics[width=1.0\linewidth]{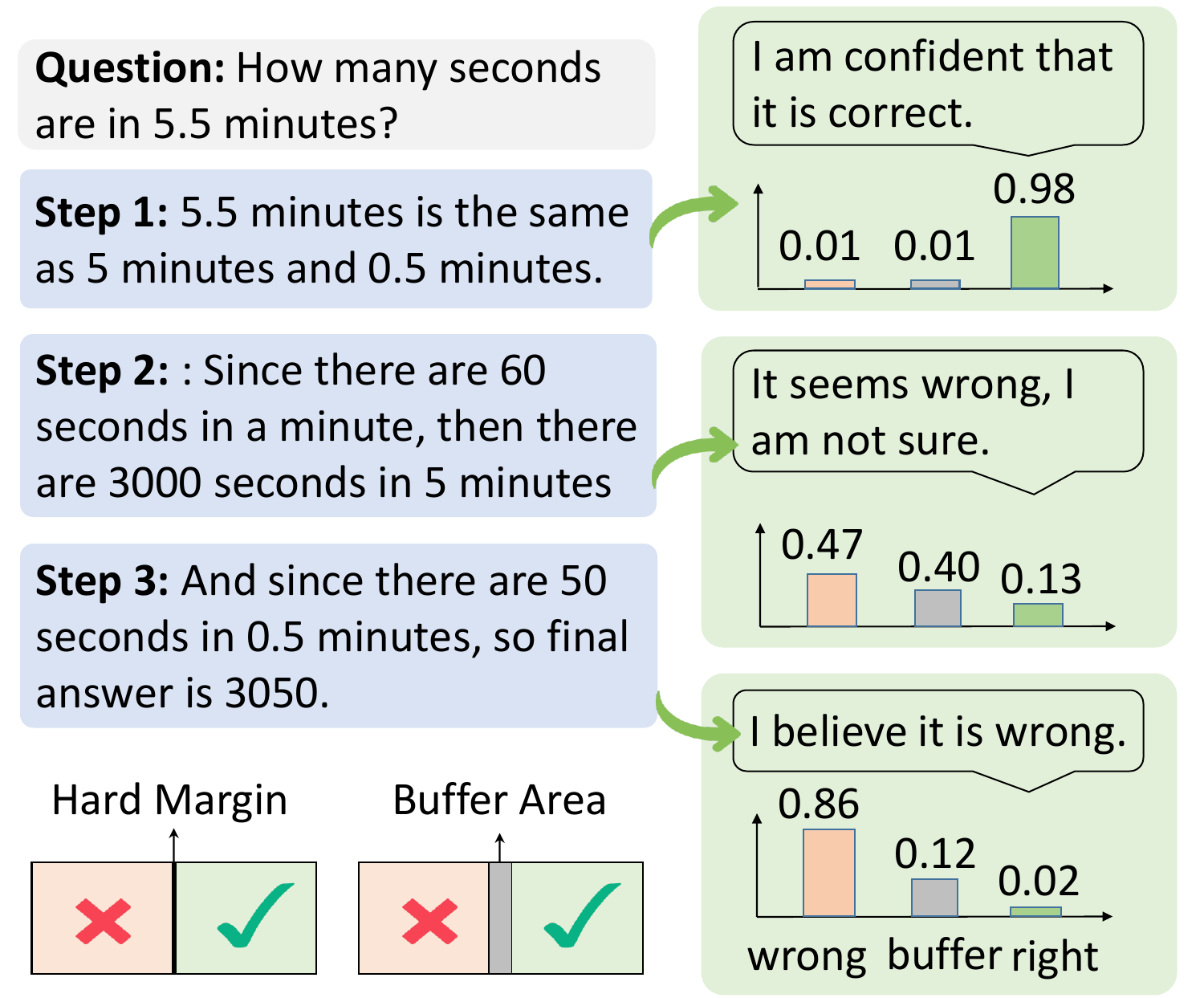}
    \caption{
        FreePRM introduces buffer area between wrong and right, along with predicting 
        probabilities of right and wrong, it also predicts buffer probability. It reflects 
        the ambiguity of uncertain reasoning steps, and helps absorb noise from unreliable pseudo-labels.
    }
    \vspace{-20pt} 
    \label{fig:intro}
\end{wrapfigure}

Unlike Outcome Reward Models (ORMs) \cite{abs-2110-14168}, which evaluate only the final result, 
Process Reward Models (PRMs) provide fine-grained feedback at each reasoning step. 
By capturing the value of intermediate steps, PRMs offer deeper insights into how individual 
steps contribute to the overall goal. This capability has made PRMs essential for 
tasks requiring complex, multi-step reasoning \cite{LightmanKBEBLLS24,WangLSXDLCWS24,abs-2408-03314}.

Despite the advantages, training PRMs remains challenging due to the high cost of annotating 
intermediate reasoning steps. While automated annotation methods, such as Monte Carlo estimation 
\cite{WangLSXDLCWS24,WangLWLH0S24} and binary search \cite{abs-2406-06592} have been proposed, 
they often demand significant 
computational resources \cite{WangLSXDLCWS24, LuD0CDFG24} and may produce noisy or unreliable labels, 
which can degrade model performance. To solve this issue, recent work \cite{YuGW24} frames PRM 
as a value estimation task, optimizing based only on the final outcome correctness. 
However, this approach overlooks the noise introduced by relying solely on final 
outcomes for supervision. These limitations lead us to the following research question:

{\centering \textit{Can we train PRM as normal classification task without annotated process labels?} \par}

Our answer is Yes. We propose {\bf FreePRM}, a framework that trains PRMs using 
only final outcome labels as weak supervision. As illustrated in Figure \ref{fig:intro} , 
Traditional binary classification methods assume that each process label is either correct or incorrect, 
with no gradation in between. We refer to this as hard margin. However, many reasoning steps exhibit 
ambiguity, making it difficult to determine their impact on the correctness of the subsequent reasoning path. 
To address this issue, a labeling approach was adopted in which each step is labeled as positive, 
negative, or neutral during manual annotation~\cite{LightmanKBEBLLS24}.
However, the neutral label is not meaningfully utilized, and is ultimately treated as either positive or negative.

Inspired by above idea, FreePRM first generates pseudo process labels by assuming that all reasoning steps 
are correct if the final outcome is correct (and incorrect otherwise). However, this approach inevitably 
introduces label noise, as a correct final answer does not guarantee the correctness of all intermediate steps. 
To mitigate the impact of such noisy labels, we introduce \textit{buffer probability} mechanism, which serves two purposes: 
(1) it represents a neutral stance toward uncertain steps, (2) it acts as a buffer that absorbs the 
influence of label inaccuracies. By extending conventional binary step prediction with this additional 
buffer state, our model can dynamically adjust its confidence in the pseudo labels and better handle 
ambiguous cases where step correctness cannot be reliably inferred.

We summarize our contributions as follows:

\begin{itemize}[itemsep=2pt,topsep=0pt,parsep=0pt]  
    \item We propose FreePRM, a novel framework for training PRMs using only binary outcome labels as weak 
    supervision. By introducing buffer probability, FreePRM effectively mitigates 
    label noise, enabling effective learning without costly step-level annotations.
    \item Empirical results on ProcessBench and mathematical reasoning tasks demonstrate that FreePRM significantly 
    outperforms PRMs trained on fully labeled data, achieving an F1 score of 53.0\%, an 24.1\% improvement 
    over the fully supervised baseline (28.9\%).
    \item We present a new perspective on PRMs by demonstrating that high-quality process evaluation can be achieved 
    without dense annotations, opening the door to more scalable and practical methods for training fine-grained PRMs.
\end{itemize}

\section{Methodology}
\subsection{FreePRM}

Our approach tackles the challenge of missing intermediate step annotations by generating pseudo process labels 
based on the correctness of final outcome and incorporating some other mechanisms to handle the resultant label noise.

\textbf{Pseudo Label Generation.} Given a reasoning trajectory $ \tau = (s_1, s_2, ..., s_T) $ consisting of $ T $ steps, and a binary final outcome label $ y \in \{0, 1\} $ indicating whether the final answer is correct, FreePRM generates pseudo step-level labels $ \hat{y}_t $ as follows:
\begin{equation}
\hat{y}_t =
\begin{cases}
1, & \text{if } y = 1 \\
0, & \text{if } y = 0
\end{cases}
\quad \forall t = 1,...,T.
\end{equation}

This assumes that all steps are correct if the final result is correct, and all steps are incorrect otherwise. While simple, this labeling strategy introduces significant label noise due to the possibility of early correct steps leading to an incorrect final result or vice versa.

\textbf{Buffer Probability.} To mitigate the impact of noisy pseudo labels, we extend the conventional binary reward model with a third buffer probability. For each step $ t $, the model predicts three probabilities: $ p^r_t $ (right), $ p^w_t $ (wrong), and $ p^b_t $ (buffer). These probabilities satisfy the constraint:
\begin{equation}
p^r_t + p^w_t + p^b_t = 1.
\end{equation}
The buffer probability $ p^b_t $ allows the model to express uncertainty when the correctness of a step cannot be confidently inferred from the final outcome alone. During training, the model is encouraged to assign higher $ p^b_t $ values to ambiguous steps, thereby reducing error propagation from potentially incorrect pseudo labels. For a given pseudo label $ \hat{y}_t \in \{0,1\} $, the training objective encourages the following behavior:
\begin{equation}
\begin{cases}
p^r_t + p^b_t \to 1 & \text{if } y_t = 1, \\
p^w_t + p^b_t \to 1 & \text{if } y_t = 0.
\end{cases}
\end{equation}

\begin{figure}[ht]
    \centering
    \includegraphics[width=1.0\textwidth]{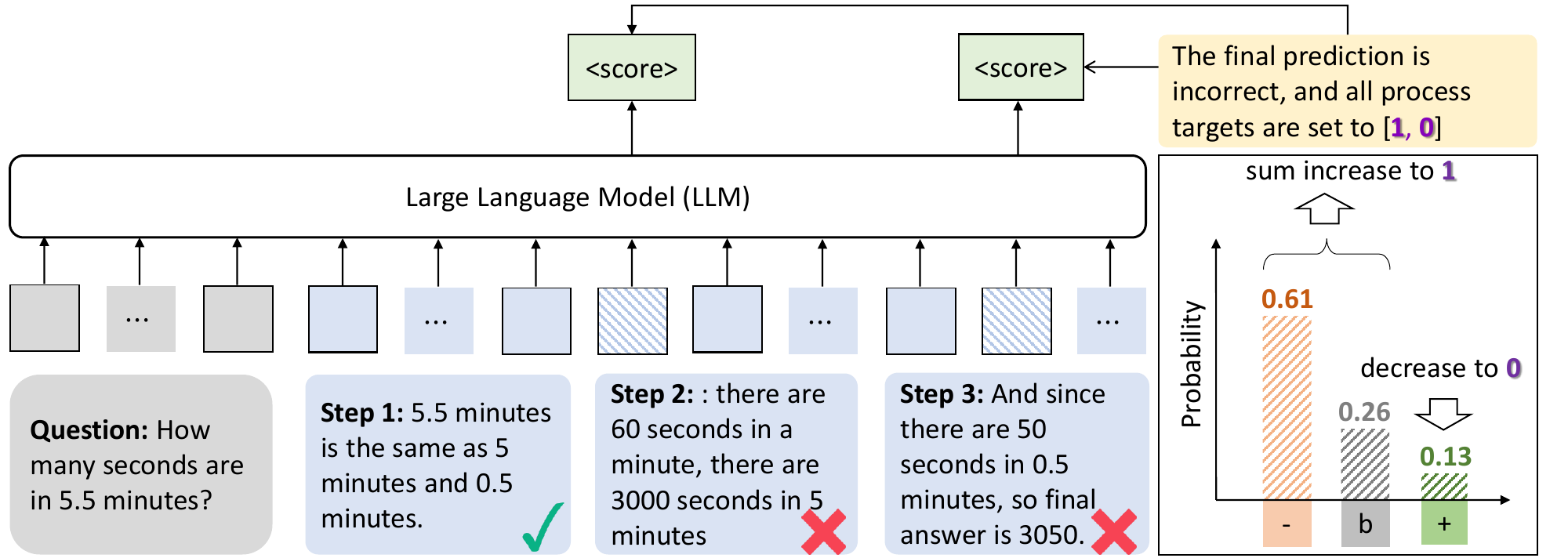}
    \caption{
        The input consists of question and its solution. The solution is divided into multiple steps, each separated by a special token:
        (``\textbackslash n\textbackslash n\textbackslash n\textbackslash n\textbackslash n''). Each step is pseudo-labeled as 
        either "+" (correct) or "-" (incorrect). FreePRM predicts three probabilities 
        for each step: wrong, buffer, and right. When a step is labeled as incorrect ("-"), the training target is set to [1, 0], 
        meaning the combined probability of wrong and buffer should sum to 1, while the probability of right is expected to be 0.
    }
    \label{fig:model_scheme}
    \end{figure}

\textbf{Random Buffer Factor.} While the buffer probability mechanism improves robustness,we observe that 
the model may collapse into assigning maximal buffer probability $ p^b_t = 1 $ at every reasoning step, 
essentially disregarding the correctness predictions $ p^r_t $ and $ p^w_t $. This occurs because $ p^b_t $ 
acts as a "safe" residual probability that trivially satisfies the learning objective without requiring 
accurate step-wise predictions.
To prevent this collapse, we introduce stochasticity into the training process. 
Instead of deterministically including $ p^b_t $, we sample a binary buffer factor $ \beta_t \in \{0, 1\} $. 
The training objective is then designed to encourage the following behavior:
\begin{equation}
\begin{cases}
    p^r_t + \beta_t p^b_t \to 1 & \text{if } y_t = 1, \\
    p^w_t + \beta_t p^b_t \to 1 & \text{if } y_t = 0.
\end{cases}
\end{equation}
where $ \beta_t \sim \text{Bernoulli}(p^b_t) $. This stochastic formulation ensures that buffer usage is 
neither guaranteed nor completely suppressed, 
promoting balanced learning dynamics and preventing model collapse (see Theorem~\ref{thm:theorem_2} 
for detailed analysis). 

\textbf{Enhanced Supervision for the Last Step.} 
We further observe that the last step in a reasoning trajectory exhibits stronge correlation with the correctness of final outcome. 
Therefore, for the last step $ t=T $, we remove the buffer probability from the 
prediction and enforce stronger supervision. To achieve this, 
we introduce a per-step weight $ \alpha_t \geq 1 $. For all steps except the last one, we set $ \alpha_t = 1 $, but for the last step, we use $ \alpha_T > 1 $ 
to amplify the loss contribution of the final step and encourage more confident predictions. The final training loss is:
\begin{equation}
\mathcal{L}(\tau) = -\frac{1}{T} \sum_{t=1}^{T} \alpha_t \left[ \hat{y}_t \log(p^r_t + \beta_t p^b_t) + (1 - \hat{y}_t) \log(p^w_t + \beta_t p^b_t) \right].
\end{equation}
This formulation integrates strong supervision at the final step while maintaining flexibility and robustness across earlier steps through the buffer probability and random buffer factor.

\subsection{Theory Analysis}
\label{sec:theory}
In this section, we analyze the robustness and stability of FreePRM through two theoretical results that demonstrate its effectiveness in mitigating label noise and preventing model collapse.
Existing works \cite{LightmanKBEBLLS24, WangLSXDLCWS24, abs-2402-03300, abs-2406-06592} normally formulate PRMs training as a classification problem and optimize the model using the cross-entropy loss:

\begin{equation}
\mathcal{L}^{\text{CE}}(\tau) = -\frac{1}{T} \sum_{t=1}^{T} \left[ \hat{y}_t \log(p^r_t) + (1 - \hat{y}_t) \log(1 - p^r_t) \right].
\end{equation}

We now establish the relationship between this cross-entropy loss and our proposed FreePRM loss.

{\bf Theorem 2.1.} 
\textit{Consider the expected FreePRM loss $\mathbb{E}[\mathcal{L}_t]$ and the cross-entropy loss $\mathcal{L}_t^{\text{CE}}$ at step $t$, under the constraint $p^r_t + p^w_t + p^b_t = 1$. Assuming $\hat{y}_t = 1$ (the case for $\hat{y}_t = 0$ follows similarly), the gradient norms with respect to $p^r_t$ satisfy:}

\begin{equation}
\left\|\frac{\partial \mathbb{E}[\mathcal{L}_t]}{\partial p^r_t}\right\| \leq \left\|\frac{\partial \mathcal{L}_t^{\text{CE}}}{\partial p^r_t} \right\| - (p_t^b)^2.
\end{equation}

\textit{Proof.}
We begin by computing the gradients of both the cross-entropy loss and the expected FreePRM loss with respect to $p^r_t$ (derived in the Appendix \ref{sec:gradient}):

\begin{equation}
\frac{\partial \mathcal{L}_t^{\text{CE}}}{\partial p^r_t} = -\frac{1}{p^r_t}, \quad
\frac{\partial \mathbb{E}[\mathcal{L}_t]}{\partial p^r_t} = -\left( \frac{p^b_t}{p^r_t + p^b_t} + \frac{1 - p^b_t}{p^r_t} \right).
\end{equation}

Taking the difference of their norms:

\begin{equation}
\left\|\frac{\partial \mathcal{L}_t^{\text{CE}}}{\partial p^r_t}\right\| - \left\|\frac{\partial \mathbb{E}[\mathcal{L}_t]}{\partial p^r_t}\right\|
= \frac{1}{p^r_t} - \frac{p^b_t}{p^r_t + p^b_t} - \frac{1 - p^b_t}{p^r_t}
= \frac{(p^b_t)^2}{p^r_t(p^r_t + p^b_t)}.
\end{equation}

From the constraint $p^r_t + p^b_t + p^w_t = 1$ and noting that $p^r_t \leq 1$, therefore
$p^r_t(p^r_t + p^b_t) \leq 1$, using this inequality, we get:

\begin{equation}
\frac{(p^b_t)^2}{T p^r_t(p^r_t + p^b_t)} \geq (p^b_t)^2.
\end{equation}

Finally, rearranging terms yields the desired result:

\begin{equation}
\left\|\frac{\partial \mathbb{E}[\mathcal{L}_t]}{\partial p^r_t}\right\| \leq \left\|\frac{\partial \mathcal{L}_t^{\text{CE}}}{\partial p^r_t}\right\| - (p^b_t)^2.
\end{equation}

The above inequality indicates that buffer probability $p^b_t$ introduces a \textit{gradient regularization},
which scales quadratically with $p^b_t$, meaning higher buffer confidence (uncertainty) induces stronger gradient suppression.
Thus it prevents the model from overfitting to noisy pseudo-labels.

\paragraph{\textbf{Theorem 2.2.}} \label{thm:theorem_2}  
\textit{With $\beta_t$ following Bernoulli distribution parameterized by $p^b_t$, the all-buffer solution where $p^b_t \to 1$ is not a fixed point in the optimization dynamics.}

\textit{Proof.}  
We begin by computing the gradient of the expected loss $ \mathbb{E}[\mathcal{L}_t] $, with respect to $ p^b_t $:

\begin{equation}
\frac{\partial \mathbb{E}[\mathcal{L}_t]}{\partial p^b_t} = -\left[ \log\left(\frac{p^r_t + p^b_t}{p^r_t}\right) + \frac{p^b_t}{p^r_t + p^b_t} \right].
\end{equation}

Now suppose $ p^b_t \to 1 $. Under the constraint $ p^r_t + p^w_t + p^b_t = 1 $, this implies $ p^r_t \to 0 $. Let $ p^r_t = \epsilon $ and $ p^b_t = 1 - \epsilon $, where $ \epsilon \to 0 $.
For the frist term:
\begin{equation}
\log\left(\frac{p^r_t + p^b_t}{p^r_t}\right) = \log\left(\frac{1}{\epsilon}\right) \to \infty.
\end{equation}

For the seconde term:
\begin{equation}
\frac{p^b_t}{p^r_t + p^b_t} = \frac{1 - \epsilon}{1} \to 1.
\end{equation}

Combining both terms yields:

\begin{equation}
\frac{\partial \mathbb{E}[\mathcal{L}_t]}{\partial p^b_t} \approx -(\infty + 1) \to -\infty.
\end{equation}

This infinitely negative gradient pushes $ p^b_t $ away from 1, making the configuration $ p^b_t \to 1 $ 
dynamically unstable. As a result, the optimization process naturally avoids the degenerate "all-buffer" solution, 
ensuring that the model remains well-behaved and avoids collapse.

\section{Experiment}
\label{sec:experiment}

\subsection{Experiment Setup}
\textbf{Training Dataset.} We conduct our experiments using the publicly available Math-Shepherd dataset \cite{WangLSXDLCWS24}, 
which generate step-level labels automatically. These labels were created as part of the process 
supervision data generation covering the MATH and GSM8K datasets. For our purposes, we exclude the process labels and 
use only the problems along with their corresponding answers.

\textbf{Evaluation.} We evaluate our model in a public PRM benchmark ProcessBench \cite{abs-2412-06559}. The aim is to judge 
whether PRM can find the first wrong step. It divides data into two parts: samples with incorrect and correct final answers 
and then conducts harmonic mean on the accuracy of these two parts to get the final F1-score. Besides, following standard 
practice \cite{LightmanKBEBLLS24}, we evaluate PRMs with best-of-N (BoN) on MATH-500 \cite{HendrycksBKABTS21}. To study the 
generalizability of the PRMs, we test each PRM using four generation models with different levels of capabilities: 
Qwen2.5-7B, MetaMath-Mistral-7B \cite{abs-2310-06825}, Muggle-Math-13B, and Llama-3-70B-Instruct \cite{abs-2407-21783}. For each 
completion, we apply PRMs to score each step and pick the last step reward as the score for overall responses. 

\textbf{Baselines.} We consider two types of baselines: (1) Language Models as Critic, including Llama \cite{abs-2407-21783}, 
Qwen2.5 \cite{abs-2412-15115}, Qwen2.5- MATH \cite{abs-2409-12122}, Qwen2.5-Coder \cite{abs-2409-12186}. These models are promoted to judge the steps with the help of majority voting; 
(2) Open-source PRM, including Skywork \cite{skywork-prm}, EurusPRM \cite{abs-2412-01981}, Qwen2.5-PRM \cite{abs-2501-07301}, Math-Shepherd \cite{WangLSXDLCWS24} 
and RLHFlow \cite{rlhflow-prm}. These models are binary classification PRMs.

\textbf{Implementation Details.} Our base model is Qwen2.5-Math-7B-Instruct, the training configuration for our method uses 
a batch size of 16 and a learning rate of 1e-4. During evaluation, we use a temperature 
setting of 0.8. The experiments are run on 2 Nvidia A40 GPUs with
BF16 precision. The prompt utilized by the policy generator is provided in 
Appendix~\ref{sec:prompt}, and the evaluation details can be found in Appendix~\ref{thm:evaluation_method}.
\subsection{Overall Performance}


\textbf{FreePRM demonstrates strong performance on ProcessBench.} As shown in 
Table~\ref{tab:result_processbench}, our model, FreePRM-7B-Math-Shepherd, 
achieves an impressive average F1 score of 53.0\%, outperforming all critic-based 
methods, including larger models (14B, 32B, 72B) and those trained on automatically 
annotated datasets, such as Skywork-PRM-7B (42.1\%) and EurusPRM-Stage2 (31.3\%). 
Notably, compared to the supervised model Qwen2.5-Math-Shepherd, which scores 28.9\%, 
FreePRM achieves an 24.1\% improvement in average F1 score on ProcessBench. 
While there remains a performance gap compared to Qwen2.5-Math-7B-PRM800K, a model 
trained on manually labeled data, this difference is relatively small given the substantial 
annotation cost required to produce the PRM800K dataset. Importantly, our model is trained 
entirely without process supervision, making the achieved performance highly competitive 
considering the significant reduction in labeling effort.

\begin{figure}[!h]
    \centering
    \includegraphics[width=1.0\textwidth]{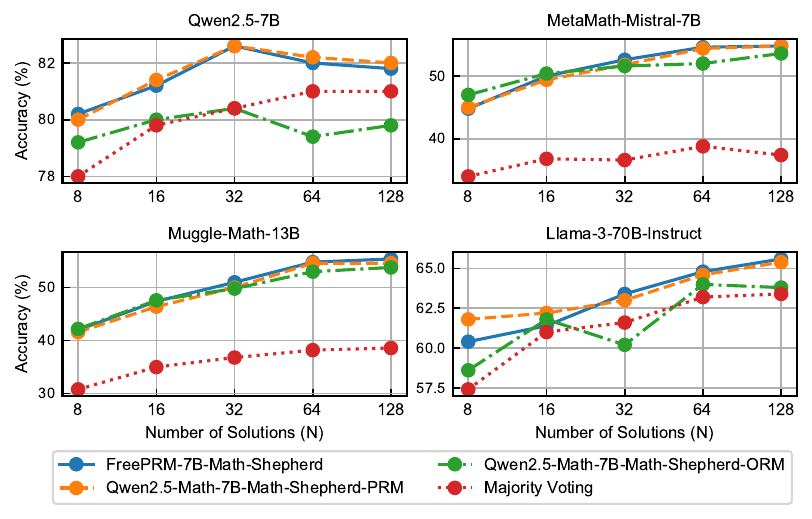}
    \caption{
        BoN results with different generation models on MATH-500. we generate 128 solutions for each problem, 
        and for each solution, we pich the final step reward as the overall score. It shows that FreePRM matches 
        or exceeds PRM which is trained on labeled process data, consistently outperforms Majority Voting and ORM 
        across  diverse model scales.
    }
    \label{fig:bon}
\end{figure}

\textbf{FreePRM matches PRMs trained on automatically labeled process data for BoN verification 
across different policy models.} We evaluate verification performance across four policy models,
with results shown in Figure~\ref{fig:bon}. FreePRM consistently outperforms baselines 
including majority voting and ORMs, particularly under low-sampling 
number. For instance, on MetaMath-Mistral-7B and Muggle-Math-13B, FreePRM achieves up to +10.4\% 
and +11.2\% improvement over majority voting at 8 samples, respectively. Compared to ORMs, 
FreePRM demonstrates superior ability in capturing intermediate reasoning quality, consistent 
with prior findings that process-based evaluation leads to more reliable assessments. Notably, 
FreePRM achieves performance comparable to or better than classification-based PRMs trained with 
BCE loss on labeled process data. These results suggest that FreePRM offers a scalable and 
step-label-free alternative for training effective PRMs without sacrificing performance.

\begin{table}[!h]
    \centering
    \caption{
        ProcessBench results reported with F1 scores. The results of FreePRM are \colorbox{cyan!10}{shaded}. 
        {\bf Avg.} indicates mean F1 across all test datasets. Top results are in {\bf bold}, and 
        runner-up results are \underline{underlined}. FreePRM outperforms baseline methods. 
        Although there remains a performance gap compared to Qwen2.5-Math-7B-PRM800K trained on high-cost manually labeled data, 
        this difference is relatively small,especially considering our model is trained entirely without process labels.
    }
    \resizebox{1.0\textwidth}{!}{
    \begin{tabular}{lcccccc}
    \toprule
    \multirow{2}{*}{\textbf{Model}} & \multirow{2}{*}{\textbf{Process Label}} & \multirow{2}{*}{\textbf{GSM8K}} & \multirow{2}{*}{\textbf{MATH}} & \multirow{2}{*}{\begin{tabular}[c]{@{}c@{}} \bf Olympiad \\ \bf Bench \end{tabular}} & \multirow{2}{*}{\begin{tabular}[c]{@{}c@{}} \bf Omni- \\ \bf MATH \end{tabular}} & \multirow{2}{*}{\textbf{Avg.}} \\
    & \\
    \midrule
    \multicolumn{7}{c}{\textit{Language Models as Critic}} \\
    \midrule
    Llama-3-8B-Instruct           & N & 13.1 & 13.8 & 4.8  & 12.6 & 11.1 \\
    Llama-3-70B-Instruct          & N & 52.2 & 22.8 & 21.2 & 20.0 & 29.1 \\
    Llama-3.1-8B-Instruct         & N & 10.9 & 5.1  & 2.8  & 1.6  & 5.1  \\
    Qwen2.5-7B-Instruct           & N & 36.5 & 36.6 & 29.7 & 27.4 & 32.6 \\
    Qwen2.5-14B-Instruct          & N & 69.3 & 53.3 & 45.0 & 41.3 & 52.2 \\
    Qwen2.5-32B-Instruct          & N & 65.6 & 53.1 & 40.0 & 38.3 & 49.3 \\
    Qwen2.5-Math-7B-Instruct      & N & 26.8 & 25.7 & 14.2 & 12.7 & 19.9 \\
    Qwen2.5-Math-72B-Instruct     & N & 65.8 & 52.1 & 32.5 & 31.7 & 45.5 \\
    Qwen2.5-Coder-7B-Instruct     & N & 14.3 & 6.5  & 4.1  & 1.8  & 6.7  \\
    Qwen2.5-Coder-14B-Instruct    & N & 50.1 & 39.9 & 34.0 & 27.3 & 37.8 \\
    \midrule
    \multicolumn{7}{c}{\textit{Process Reward Models (PRMs)}} \\
    \midrule
    Qwen2.5-Math-7B-PRM800K       & Y (manual label) & \underline{68.2} & \textbf{62.6} & \textbf{50.7} & \textbf{44.3} & \textbf{56.5} \\
    Math-Shepherd-PRM-7B          & Y & 47.9 & 29.5 & 24.8 & 23.8 & 31.5 \\
    RLHFlow-PRM-Mistral-8B        & Y & 50.4 & 33.4 & 13.8 & 15.8 & 28.4 \\
    RLHFlow-PRM-Deepseek-8B       & Y & 38.8 & 33.8 & 16.9 & 16.9 & 26.6 \\
    Skywork-PRM-7B                & Y & 70.8 & 53.6 & 22.9 & 21.0 & 42.1 \\
    EurusPRM-Stage1               & Y & 44.3 & 35.6 & 21.7 & 23.1 & 31.2 \\
    EurusPRM-Stage2               & Y & 47.3 & 35.7 & 21.2 & 20.9 & 31.3 \\
    Qwen2.5-Math-7B-Math-Shepherd-PRM & Y & 62.5 & 31.6 & 13.7 & 7.7  & 28.9 \\
    \rowcolor{cyan!10} FreePRM-7B-Math-Shepherd (ours) & N  & \textbf{74.2} & \underline{58.8} & \underline{39.0} & \underline{40.1} & \underline{53.0} \\
    \bottomrule
    \end{tabular}
    }
    \label{tab:result_processbench}
    \end{table}

\subsection{Ablation Study}

\textbf{FreePRM achieves optimal performance with full training data, while showing 
considerable effectiveness with just 20\% of the data.} Experimental results demonstrate 
that FreePRM attains its best performance when trained on 100\% of the data, achieving a 
score of 53.0\%. Notably, the model performs surprisingly well even with only 20\% of the 
training data, scoring 49.1\%. In contrast, intermediate data levels (40\% to 80\%) yield 
lower and more inconsistent performance. These findings suggest that while FreePRM can 
effectively learn reward signals from limited data, full training coverage significantly 
enhances its ability to accurately distinguish correct reasoning steps, resulting in more 
robust and reliable inference.

\begin{table}[!h]
    \centering
    \caption{Results on ProcessBench using different amounts of training data. We report the accuracies on 
    both erroneous and correct samples, along with the F1 score. 
    The F1 score is our primary metric for comparing model performance. 
    We observe that FreePRM achieves the highest performance when trained on 100\% of the data, while using 
    only 20\% of the data still yields considerable performance.
}
    \resizebox{\textwidth}{!}{
    \begin{tabular}{cccccccccccccc}
    \toprule
    \multirow{2}[2]{*}{\textbf{Percent}} & \multicolumn{3}{c}{\textbf{GSM8K}} & \multicolumn{3}{c}{\textbf{MATH}} & \multicolumn{3}{c}{\textbf{OlympiadBench}} & \multicolumn{3}{c}{\textbf{Omni-MATH}} & \multirow{2}[2]{*}{\begin{tabular}[c]{@{}c@{}} \bf Avg. \\ \bf F1 \end{tabular}} \\
    \cmidrule(lr){2-4} \cmidrule(lr){5-7} \cmidrule(lr){8-10} \cmidrule(lr){11-13}
    & Err. & Corr. & \textbf{F1} & Err. & Corr. & \textbf{F1} & Err. & Corr. & \textbf{F1} & Err. & Corr. & \textbf{F1} \\
    \midrule
    20\%       & 55.6  & 89.1    & 68.4      & 43.3  & 83.0    & 56.9      & 24.8    & 69.3    & 36.5          & 23.6    & 65.6    & 34.7        & 49.1                \\
    40\%       & 51.7  & {\bf 93.3}    & 66.5      & 33.2  & 90.9    & 48.6      & 18.5     & 82.9    & 30.2         & 12.4     & 83.4    & 21.6    & 41.7               \\
    60\%       & 59.4  & 93.3    & 72.6      & 36.4  & {\bf 93.3}   &  52.3      & 15.7     & {\bf 89.4}    & 26.8          & 15.0    & {\bf 87.6}    &  25.6       & 44.3               \\
    80\%       & 57.5  & 75.1    & 65.1     & {\bf 45.8}  & 64.3    & 53.5      & 27.4     & 40.4    & 32.6          & {\bf 28.2}     & 44.0    & 34.4        & 46.4               \\
    100\%      & {\bf 63.8}  & 88.6    & {\bf 74.2}      & 44.8  & 85.5    & {\bf 58.8}      & {\bf 27.4}    & 67.8    & {\bf 39.0}          & 27.7    & 72.6   & {\bf 40.1}        & {\bf 53.0}               \\
 \bottomrule
    \end{tabular}
    }
\end{table}


\textbf{Both random buffer factor and enhanced supervision for the last step contribute to improved performance.} 
To evaluate the effectiveness of different components in FreePRM, we conduct ablation 
studies on ProcessBench, as shown in Table~\ref{tab:result_ablation}. When neither 
component is used, the model achieves an average score of 28.5\%. Adding only the Random 
Buffer Factor leads to notable improvements across all datasets, with the average score 
increasing to 34.3\%, demonstrating its positive impact on reward modeling. significantly, 
when both components are included, the model achieves strong performance across all tasks, 
reaching 49.1\% in average score, an improvement of over 20\% compared to the baseline. 
This indicates that both components contribute to enhancing the overall effectiveness of 
FreePRM.

\textbf{Appropriate last-step weight boosts performance of FreePRM.} 
We study the effect of varying the weight assigned to the last reasoning step in the 
training objective. As shown in Table~\ref{tab:last_weight}, increasing the weight from 
1.0 to 3.0 leads to substantial improvements, with the average F1 score rising from 34.3 
to 49.1. This indicates that emphasizing the correctness of the final reasoning step 
helps the model boost performance. However, further increasing the weight beyond 3.0 
results in diminishing results and even performance degradation. These results demonstrate 
that appropriately strengthening supervision for the last step significantly enhances 
FreePRM’s ability to evaluate complex reasoning trajectory.

\begin{table}[!h]
    \centering
    \caption{Results on ProcessBench with different components show that both the Random Buffer Factor and Enhanced Supervision for the Last Step improve performance. (trained with 20\% data)}
    \resizebox{0.9\textwidth}{!}{
    \begin{tabular}{cccccccc}
    \toprule
    \multirow{2}{*}{\textbf{No.}}  & \multirow{2}{*}{\begin{tabular}[c]{@{}c@{}} \bf Random \\ \bf Buffer Factor \end{tabular}} & \multirow{2}{*}{\begin{tabular}[c]{@{}c@{}} \bf Enhanced \\ \bf Last Step \end{tabular}} & \multirow{2}{*}{\textbf{GSM8K}} & \multirow{2}{*}{\textbf{MATH}} & \multirow{2}{*}{\begin{tabular}[c]{@{}c@{}} \bf Olympiad \\ \bf Bench \end{tabular}} & \multirow{2}{*}{\begin{tabular}[c]{@{}c@{}} \bf Omni- \\ \bf MATH \end{tabular}} & \multirow{2}{*}{\textbf{Avg.}} \\
    & \\
    \midrule
    1 & \ding{55} & \ding{55} & 38.9 & 36.8 & 19.2 & 19.1 & 28.5 \\
    2 & \ding{51} & \ding{55}& 50.4 & 41.4 & 21.3 & 24.0 & 34.3 \\
    3 & \ding{51} & \ding{51}& {\bf 68.4} & {\bf 56.9} & {\bf 36.5} & {\bf 34.7} & {\bf 49.1} \\
    \bottomrule
    \end{tabular}
    }
    \label{tab:result_ablation}
    \end{table}%

\begin{table}[!h]
    \centering
    \caption{Performance of FreePRM on ProcessBench with different weight settings for the last step, appropriate weight setting lead to significant improvements. (trained with 20\% data)}
    \resizebox{\textwidth}{!}{
    \begin{tabular}{cccccccccccccc}
    \toprule
    \multirow{2}[2]{*}{\textbf{Weight}} & \multicolumn{3}{c}{\textbf{GSM8K}} & \multicolumn{3}{c}{\textbf{MATH}} & \multicolumn{3}{c}{\textbf{OlympiadBench}} & \multicolumn{3}{c}{\textbf{Omni-MATH}} & \multirow{2}[2]{*}{\begin{tabular}[c]{@{}c@{}} \bf Avg. \\ \bf F1 \end{tabular}} \\
    \cmidrule(lr){2-4} \cmidrule(lr){5-7} \cmidrule(lr){8-10} \cmidrule(lr){11-13}
    & Err. & Corr. & \textbf{F1} & Err. & Corr. & \textbf{F1} & Err. & Corr. & \textbf{F1} & Err. & Corr. & \textbf{F1} \\
    \midrule
    1.0   & 38.2  & 74.1    & 50.4      & 29.1  & 71.4    & 41.4      & 13.6    & 48.7    & 21.3          & 15.0    & 59.3    & 24.0        & 34.3                \\
    3.0   & {\bf 55.6}  & {\bf 89.1}    & {\bf 68.4}      & {\bf 43.3}  & {\bf 83.0}    & {\bf 56.9}      & {\bf 24.8}    & 69.3    & {\bf 36.5}          & {\bf 23.6}    & 65.6    & {\bf 34.7}        & {\bf 49.1}                \\
    5.0   & 38.2  & 77.2    & 51.1      & 29.1  & 82.8    & 43.1      & 15.1    & {\bf 73.2}    & 25.1          & 14.4    & {\bf 78.0}    & 24.3        & 35.9                \\
    7.0   & 35.7  & 80.8    & 49.6      & 26.9  & 79.6    & 40.2      & 11.5    & 64.9    & 19.5          & 9.1    & 73.0    & 16.2        & 31.4               \\
    \bottomrule
    \end{tabular}
    }
    \label{tab:last_weight}
\end{table}

\section{Related Work}
\textbf{Process Labels annotating.} Generating high-quality process supervision data is 
crucial for training PRMs. There are two primary approaches to annotating process labels.
The first relies on human annotation \cite{LightmanKBEBLLS24}, which yields high-quality 
labels but at a significant cost. The second approach is automated evaluation of 
reasoning step correctness. Current methods fall into two categories:(1) 
Backward-propagation-based methods that infer step correctness from final outcomes, 
including Monte Carlo estimation \cite{WangLSXDLCWS24, abs-2406-06592, ChenL0024a}, 
binary search \cite{abs-2406-06592}, progressive ORM labeling \cite{XiCHJZHDLGWGSFZ24}, and credit 
assignment techniques \cite{abs-2406-14283, abs-2502-01456, abs-2412-01981}. 
(2) Critic-based methods that use LLMs as critics (often referred to as LLM-as-a-judge) 
to directly assess the correctness of reasoning steps \cite{abs-2408-15240, abs-2406-14024, Xia0LWL25}. 
Combining Monte Carlo estimation with LLM-as-a-judge was further proposed to improve labeling 
accuracy \cite{abs-2501-07301}. However, automated methods generally require significant 
computational resources and may produce noisy or unreliable labels, which can degrade 
model performance.

\textbf{Process Reward Model Training.} Most existing works 
\cite{LightmanKBEBLLS24, WangLSXDLCWS24, abs-2402-03300, abs-2406-06592}
formulate PRM as a classification task, where the reward for each reasoning step is 
modeled as the probability of its correctness, and optimize the model using cross-entropy 
loss.  Building on this,  PQM \cite{abs-2410-11287} reframes PRM as a Q-value 
ranking problem. It introduces a novel comparative loss to optimize the relative rankings 
of Q-values, enhancing the model's ability to capture the sequential dependencies and 
complex dynamics of multi-step reasoning. However, both classification-based and Q-value 
ranking methods rely heavily on high-quality process labels.  To reduce dependence on 
costly labeled data, OVM \cite{YuGW24} formulates PRM as a value estimation task, using MSE loss 
to predict the potential value of an incomplete reasoning path, based solely on the 
correctness of the final outcome. While this avoids explicit labeling, it overlooks 
the label noise inherent in deriving supervision signals purely from final outcomes, 
potentially degrading training stability and performance.

\section{Conclusion and Limitations}\label{sec:conclusion}
Training PRMs without costly step-wise annotations 
presents a significant challenge, as existing methods rely on resource-intensive 
labeling or step-labels derived solely from final outcomes. 
We address this with FreePRM, a framework that trains PRMs using only binary 
outcome labels by introducing a buffer probability mechanism to mitigate label 
noise. This buffer dynamically adjusts confidence in pseudo-labels, representing 
uncertainty for ambiguous steps while absorbing inaccuracies. Empirical results 
on ProcessBench and mathematical reasoning tasks demonstrate FreePRM’s 
effectiveness, achieving a 53.0\% F1 score, a 24.1\% improvement over fully 
supervised baselines, without requiring annotated process labels. Our work 
challenges the necessity of dense annotations for PRMs, offering a scalable, 
noise-robust alternative that advances practical deployment of fine-grained 
process evaluation in complex reasoning tasks.

One limitation of FreePRM is the performance gap compared to PRM models trained 
on high-quality, manually annotated datasets. It would be interesting to explore how 
FreePRM can be combined with automated data annotation techniques to bridge this 
performance gap. We consider this an exciting direction for future research.

\bibliography{references}
\bibliographystyle{plain}

\newpage
\appendix


\section{Derivations and Proofs}
\label{sec:gradient}

The FreePRM loss function is defined as:

\begin{equation}
\mathcal{L}(\tau) = -\frac{1}{T} \sum_{t=1}^{T} \alpha_t \left[ \hat{y}_t \log(p^r_t + \beta_t p^b_t) + (1 - \hat{y}_t) \log(p^w_t + \beta_t p^b_t) \right].
\end{equation}

where $\beta_t \sim \text{Bernoulli}(p^b_t)$, meaning $\beta_t = 1$ with probability $p^b_t$ and $\beta_t = 0$ otherwise;
$\hat{y}_t \in \{0,1\}$ is the fixed ground-truth label at time step $t$;
The normalization constraint is: $p^r_t + p^w_t + p^b_t = 1$.

Given that $\beta_t$ is a Bernoulli random variable with:
\begin{equation}
\mathbb{P}(\beta_t = 1) = p^b_t,\quad \mathbb{P}(\beta_t = 0) = 1 - p^b_t/
\end{equation}

we compute the expected value of the loss at time step $t$, denoted $\mathcal{L}_t$.
For clarity, we focus on the case where $ y_i = 1 $, the case where $ y_i = 0 $ follows similarly. Besides, we take the normal step where $\alpha_t = 1$, the loss simplifies to:

\begin{equation}
\mathbb{E}_{\beta_t}[\mathcal{L}_t] = -\mathbb{E}_{\beta_t} \left[ \hat{y}_t \log(p^r_t + \beta_t p^b_t) \right].
\end{equation}

Computing the expectation explicitly gives:

\begin{equation}
\mathbb{E}_{\beta_t}[\log(p^r_t + \beta_t p^b_t)] =-\left( p^b_t \cdot \log(p^r_t + p^b_t) + (1 - p^b_t) \cdot \log(p^r_t) \right).
\end{equation}

Taking the partial derivative of $\mathbb{E}[\mathcal{L}_t]$ with respect to $p^r_t$, we obtain:

\begin{equation}
\boxed{
\frac{\partial \mathbb{E}[\mathcal{L}_t]}{\partial p^r_t} = -\left(\frac{p^b_t}{p^r_t + p^b_t} + \frac{1 - p^b_t}{p^r_t}\right).
}
\end{equation}

Similarly, taking the partial derivative with respect to $p^b_t$, we have:

\begin{equation}
\boxed{
\frac{\partial \mathbb{E}[\mathcal{L}_t]}{\partial p^b_t} = -\left[ \log\left(\frac{p^r_t + p^b_t}{p^r_t}\right) + \frac{p^b_t}{p^r_t + p^b_t} \right].
}
\end{equation}

\section{Prompt}
\label{sec:prompt}

The prompt used for the policy generator is shown below.

\begin{tcolorbox}[title=Prompt for policy generator, label={tab:rationale_prompt}, breakable, width=\textwidth,
fonttitle=\bfseries
]
\textbf{[System]:} \\
Please reason step by step, and put your final answer within boxed\{\}. \\

\textbf{[User]:} \\
The following is the math problem: \\

[Math Problem] \\

\{problem\} \\

Let's think step by step and output the final answer within boxed\{\}. \\
\end{tcolorbox}

\section{Additional Results}

We provide full results of ProcessBench in Table~\ref{tab:all_processbench}.

\begin{table}[H]
\centering
\caption{Full results of critic models and PRMs on ProcessBench.}
\resizebox{\textwidth}{!}{
\begin{tabular}{lccccccccccccc}
\toprule
\multirow{2}[2]{*}{\textbf{Model}} & \multicolumn{3}{c}{\textbf{GSM8K}} & \multicolumn{3}{c}{\textbf{MATH}} & \multicolumn{3}{c}{\textbf{OlympiadBench}} & \multicolumn{3}{c}{\textbf{Omni-MATH}} & \multirow{2}[2]{*}{\begin{tabular}[c]{@{}c@{}} \bf Avg. \\ \bf F1 \end{tabular}} \\
\cmidrule(lr){2-4} \cmidrule(lr){5-7} \cmidrule(lr){8-10} \cmidrule(lr){11-13}
& Err. & Corr. & \textbf{F1} & Err. & Corr. & \textbf{F1} & Err. & Corr. & \textbf{F1} & Err. & Corr. & \textbf{F1} \\
\midrule
\multicolumn{14}{c}{\textit{Language Models as Critic}} \\
\midrule
Llama-3-8B-Instruct           & 42.5 & 7.8   & 13.1 & 28.6 & 9.1  & 13.8 & 27.1 & 2.7  & 4.8  & 26.1 & 8.3  & 12.6 & 11.1 \\
Llama-3-70B-Instruct          & 35.7 & 96.9  & 52.2 & 13.0 & 93.3 & 22.8 & 12.0 & 92.0 & 21.2 & 11.2 & 91.7 & 20.0 & 29.1 \\
Llama-3.1-8B-Instruct         & 44.4 & 6.2   & 10.9 & 41.9 & 2.7  & 5.1  & 32.4 & 1.5  & 2.8  & 32.0 & 0.8  & 1.6  & 5.1  \\
Qwen2.5-7B-Instruct           & 40.6 & 33.2  & 36.5 & 30.8 & 45.1 & 36.6 & 26.5 & 33.9 & 29.7 & 26.2 & 28.6 & 27.4 & 32.6 \\
Qwen2.5-14B-Instruct          & 54.6 & 94.8  & 69.3 & 38.4 & 87.4 & 53.3 & 31.5 & 78.8 & 45.0 & 28.3 & 76.3 & 41.3 & 52.2 \\
Qwen2.5-32B-Instruct          & 49.3 & 97.9  & 65.6 & 36.7 & 95.8 & 53.1 & 25.3 & 95.9 & 40.0 & 24.1 & 92.5 & 38.3 & 49.3 \\
Qwen2.5-Math-7B-Instruct      & 15.5 & 100.0 & 26.8 & 14.8 & 96.8 & 25.7 & 7.7  & 91.7 & 14.2 & 6.9  & 88.0 & 12.7 & 19.9 \\
Qwen2.5-Math-72B-Instruct     & 49.8 & 96.9  & 65.8 & 36.0 & 94.3 & 52.1 & 19.5 & 97.3 & 32.5 & 19.0 & 96.3 & 31.7 & 45.5 \\
Qwen2.5-Coder-7B-Instruct     & 7.7  & 100.0 & 14.3 & 3.4  & 98.3 & 6.5  & 2.1  & 99.1 & 4.1  & 0.9  & 98.3 & 1.8  & 6.7  \\
Qwen2.5-Coder-14B-Instruct    & 33.8 & 96.4  & 50.1 & 25.4 & 92.4 & 39.9 & 20.7 & 94.1 & 34.0 & 15.9 & 94.2 & 27.3 & 37.8 \\
\midrule
\multicolumn{14}{c}{\textit{Process Reward Models (PRMs)}} \\
\midrule
Qwen2.5-Math-7B-PRM800K       & 53.1 & 95.3  & 68.2 & 48.0 & 90.1 & 62.6 & 35.7 & 87.3 & 50.7 & 29.8 & 86.1 & 44.3 & 56.5 \\
Math-Shepherd-PRM-7B          & 32.4 & 91.7  & 47.9 & 18.0 & 82.0 & 29.5 & 15.0 & 71.1 & 24.8 & 14.2 & 73.0 & 23.8 & 31.5 \\
RLHFlow-PRM-Mistral-8B        & 33.8 & 99.0  & 50.4 & 21.7 & 72.2 & 33.4 & 8.2  & 43.1 & 13.8 & 9.6  & 45.2 & 15.8 & 28.4 \\
RLHFlow-PRM-Deepseek-8B       & 24.2 & 98.4  & 38.8 & 21.4 & 80.0 & 33.8 & 10.1 & 51.0 & 16.9 & 10.9 & 51.9 & 16.9 & 26.6 \\
Skywork-PRM-7B                & 61.8 & 82.9  & 70.8 & 43.8 & 62.2 & 53.6 & 17.9 & 31.9 & 22.9 & 14.0 & 41.9 & 21.0 & 42.1 \\
EurusPRM-Stage1               & 46.9 & 42.0  & 44.3 & 33.3 & 38.2 & 35.6 & 23.9 & 19.8 & 21.7 & 21.9 & 24.5 & 23.1 & 31.2 \\
EurusPRM-Stage2               & 51.2 & 44.0  & 47.3 & 36.4 & 35.0 & 35.7 & 25.7 & 18.0 & 21.2 & 23.1 & 19.1 & 20.9 & 31.3 \\
Qwen2.5-Math-7B-Math-Shepherd-PRM & 46.4 & 95.9  & 62.5 & 18.9 & 96.6 & 31.6 & 7.4  & 93.8 & 13.7 & 4.0  & 95.0 & 7.7  & 28.9 \\
FreePRM-7B-Math-Shepherd (ours)   & 63.8 & 88.6  & 74.2 & 44.8 & 85.5 & 58.8 & 27.4 & 67.8 & 39.0 & 27.7 & 72.6 &40.1  & 53.0 \\
\bottomrule
\end{tabular}
}
\label{tab:all_processbench}
\end{table}

\begin{figure}[!h]
    \centering
    \includegraphics[width=1.0\textwidth]{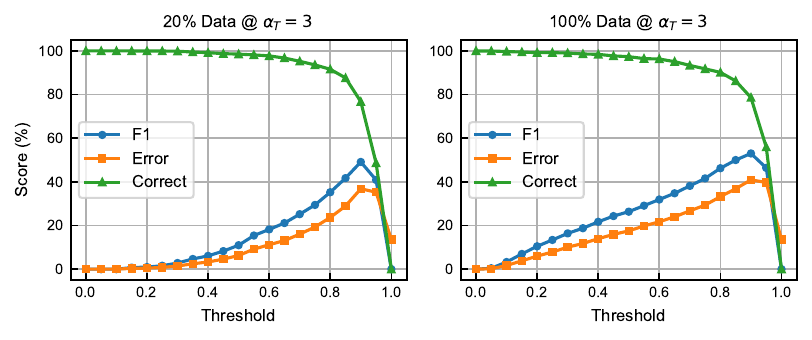}
    \caption{
        Influence of right score threshold selection on overall performance. As the threshold for the right score 
        increases, the correct accuracy gradually decreases. Conversely, both the error accuracy and the F1 score 
        initially increase but then sharply decline. The optimal F1 performance is achieved when the threshold is 
        set to 0.9.}
    \label{fig:thres}
    \end{figure}

\paragraph{\textbf{ Evaluation of FreePRM.}} \label{thm:evaluation_method}  
After training the FreePRM model, we use a policy generator to generate problem solutions. Each step in these solutions is then evaluated by FreePRM, which assigns three scores: right, wrong, and buffer. These scores sum to one, and only the right score is used for error detection in PressBench and for BoN score comparisons.
In the PressBench evaluation, we first define a threshold for the right score. As we iterate through each step of the solution, any step with a right score below this threshold is flagged as an error.
Figure \ref{fig:thres} shows the impact of different threshold settings on the final performance metrics.

\section{Case studies}
\label{apd:case-study}
We show more qualatative example in Table~\ref{tab:apd-case-1}. 
The blue values represent the predicted right probabilities provided by FreePRM.
the probability is low when encountering the wrong step.
\begin{table}[!h]
    \centering
    \caption{Case study on ProcessBench. Solutions are sampled from Llama-3.1-8B-Instruct. 
    The blue value represents right score predicted by FreePRM.}
    \begin{tabularx}{1.0\linewidth}{X}
    \toprule
    \textbf{Question:} 
    \begin{tcolorbox}[colback=nmgray, colframe=white, width=13cm, arc=0mm]
        The Coventry School's European debate club has 6 German delegates, 5 French delegates, 
        and 3 Italian delegates.  In how many ways can these 14 delegates sit in a row of 14 chairs, 
        if each country's delegates insist on all sitting next to each other? (first wrong step: 3)
        \end{tcolorbox} \\

    \midrule
    \textbf{ Step 1:} \begin{tcolorbox}[colback=cyan!10, colframe=white, width=13cm, arc=0mm]
        To find the total number of ways the delegates can sit in a row of 14 chairs with each 
        country's delegates sitting together, we can break down the problem into smaller steps. 
        First, consider the three groups of delegates as single units. Since there are three 
        countries, we have 3 units (groups) that need to be arranged in a row. The number of ways 
        to arrange these 3 units is 3!. This is because for the first position, we have 3 choices, 
        for the second position, we have 2 choices left, and for the last position, we have only 
        1 choice left. 3! = 3 * 2 * 1 = 6. 
        \newline \textcolor{blue}{(0.97)}
        \end{tcolorbox} \\

    \textbf{ Step 2:} \begin{tcolorbox}[colback=cyan!10, colframe=white, width=13cm, arc=0mm]
        Next, let's consider the internal arrangement of each group. For the 6 German delegates, 
        they can be arranged in 6! ways. For the 5 French delegates, they can be arranged in 5! 
        ways. And for the 3 Italian delegates, they can be arranged in 3! ways. 
        6! = 6 * 5 * 4 * 3 * 2 * 1 = 720, 5! = 5 * 4 * 3 * 2 * 1 = 120, 
        and 3! = 3 * 2 * 1 = 6. 
        \newline \textcolor{blue}{(0.99)}
        \end{tcolorbox} \\

    \textbf{ Step 3 (wrong step):} \begin{tcolorbox}[colback=pink!20, colframe=white, width=13cm, arc=0mm]
        To find the total number of arrangements, we multiply the number of ways to arrange 
        the units (Step 2) by the number of ways to arrange each group internally (Step 3). 
        Total number of arrangements = (number of ways to arrange units) * 
        (number of ways to arrange Germans) * (number of ways to arrange French) *
        (number of ways to arrange Italians) = 6 * 720 * 120 * 6 = 6 * 518400 = 3104400.
        \newline \textcolor{blue}{(0.11)}
        \end{tcolorbox} \\

    \textbf{ Step 4 (wrong step):} \begin{tcolorbox}[colback=pink!20, colframe=white, width=13cm, arc=0mm]
        Therefore, the final answer is: \boxed{3104400}
        \newline \textcolor{blue}{(0.19)}
        \end{tcolorbox} \\
    
    \bottomrule
    \end{tabularx} 
    \label{tab:apd-case-1}
\end{table}

\section{Societal Impacts}
\label{app:societal-impacts}

FreePRM, a method for training process reward models with minimal supervision, offers 
significant opportunities. It democratizes AI by lowering the barrier to developing reliable process-level systems, 
as it reduces the need for expensive, step-wise labeled data. This can empower 
researchers and organizations with limited resources to build robust models in 
critical areas such as education, healthcare, and scientific workflows. 
Additionally, FreePRM enhances transparency in AI decision-making by inferring 
step-wise correctness from outcomes, which is particularly valuable in sequential 
tasks like tutoring systems and medical diagnostics, fostering trust and accountability.

\clearpage
\newpage
\section*{NeurIPS Paper Checklist}

\begin{enumerate}

\item {\bf Claims}
    \item[] Question: Do the main claims made in the abstract and introduction accurately reflect the paper's contributions and scope?
    \item[] Answer: \answerYes{} 
    \item[] Justification: Our main claim matches our theoretical and experimental results in Section~\ref{sec:theory} and Section~\ref{sec:experiment}.

\item {\bf Limitations}
    \item[] Question: Does the paper discuss the limitations of the work performed by the authors?
    \item[] Answer: \answerYes{} 
    \item[] Justification: Please see Section~\ref{sec:conclusion} for the discussion of limitations.

\item {\bf Theory assumptions and proofs}
    \item[] Question: For each theoretical result, does the paper provide the full set of assumptions and a complete (and correct) proof?
    \item[] Answer: \answerYes{} 
    \item[] Justification: Please refer to Section~\ref{sec:theory} for our assumptions and a complete (and correct) proof.

    \item {\bf Experimental result reproducibility}
    \item[] Question: Does the paper fully disclose all the information needed to reproduce the main experimental results of the paper to the extent that it affects the main claims and/or conclusions of the paper (regardless of whether the code and data are provided or not)?
    \item[] Answer: \answerYes{} 
    \item[] Justification: We provide information needed to reproduce the main experimental result and the link which contains the code and dataset to reproduce our results.

\item {\bf Open access to data and code}
    \item[] Question: Does the paper provide open access to the data and code, with sufficient instructions to faithfully reproduce the main experimental results, as described in supplemental material?
    \item[] Answer: \answerYes{}  
    \item[] Justification: The code of FreePRM is available

\item {\bf Experimental setting/details}
    \item[] Question: Does the paper specify all the training and test details (e.g., data splits, hyperparameters, how they were chosen, type of optimizer, etc.) necessary to understand the results?
    \item[] Answer: \answerYes{} 
    \item[] Justification: We provide all the training and test details in Section~\ref{sec:experiment}.

\item {\bf Experiment statistical significance}
    \item[] Question: Does the paper report error bars suitably and correctly defined or other appropriate information about the statistical significance of the experiments?
    \item[] Answer: \answerYes{} 
    \item[] Justification: We provide the experiment results that support the main claims of the paper.

\item {\bf Experiments compute resources}
    \item[] Question: For each experiment, does the paper provide sufficient information on the computer resources (type of compute workers, memory, time of execution) needed to reproduce the experiments?
    \item[] Answer: \answerYes{} 
    \item[] Justification: We provide the computer resources in Section~\ref{sec:experiment}.
    
\item {\bf Code of ethics}
    \item[] Question: Does the research conducted in the paper conform, in every respect, with the NeurIPS Code of Ethics \url{https://neurips.cc/public/EthicsGuidelines}?
    \item[] Answer: \answerYes{} 
    \item[] Justification: We make sure to preserve anonymity and conform NeurIPS Code of Ethics.

\item {\bf Broader impacts}
    \item[] Question: Does the paper discuss both potential positive societal impacts and negative societal impacts of the work performed?
    \item[] Answer: \answerYes{} 
    \item[] Justification:Please see Appendix~\ref{app:societal-impacts} for broader impacts.
    
\item {\bf Safeguards}
    \item[] Question: Does the paper describe safeguards that have been put in place for responsible release of data or models that have a high risk for misuse (e.g., pretrained language models, image generators, or scraped datasets)?
    \item[] Answer: \answerNA{} 
    \item[] Justification: The paper poses no such risks.

\item {\bf Licenses for existing assets}
    \item[] Question: Are the creators or original owners of assets (e.g., code, data, models), used in the paper, properly credited and are the license and terms of use explicitly mentioned and properly respected?
    \item[] Answer: \answerNA{} 
    \item[] Justification: The paper does not use existing assets.

\item {\bf New assets}
    \item[] Question: Are new assets introduced in the paper well documented and is the documentation provided alongside the assets?
    \item[] Answer: \answerNA{} 
    \item[] Justification: The paper does not release new assets.

\item {\bf Crowdsourcing and research with human subjects}
    \item[] Question: For crowdsourcing experiments and research with human subjects, does the paper include the full text of instructions given to participants and screenshots, if applicable, as well as details about compensation (if any)? 
    \item[] Answer: \answerNA{} 
    \item[] Justification: The paper does not involve crowdsourcing nor research with human subjects.

\item {\bf Institutional review board (IRB) approvals or equivalent for research with human subjects}
    \item[] Question: Does the paper describe potential risks incurred by study participants, whether such risks were disclosed to the subjects, and whether Institutional Review Board (IRB) approvals (or an equivalent approval/review based on the requirements of your country or institution) were obtained?
    \item[] Answer: \answerNA{} 
    \item[] Justification: The paper does not involve crowdsourcing nor research with human subjects.

\item {\bf Declaration of LLM usage}
    \item[] Question: Does the paper describe the usage of LLMs if it is an important, original, or non-standard component of the core methods in this research? Note that if the LLM is used only for writing, editing, or formatting purposes and does not impact the core methodology, scientific rigorousness, or originality of the research, declaration is not required.
    \item[] Answer: \answerNA{} 
    \item[] Justification: The core method development in this research does not involve LLMs as any important, original, or non-standard components.

\end{enumerate}

\end{document}